\def\year{2021}\relax
\documentclass[letterpaper]{article} 
\usepackage{aaai21}  
\usepackage{times}  
\usepackage{helvet} 
\usepackage{courier}  
\usepackage[hyphens]{url}  
\usepackage{graphicx} 
\usepackage{natbib}  
\usepackage{caption} 
\usepackage{multirow}
\usepackage{subfigure}
\usepackage{epsfig}
\usepackage{amsmath}
\usepackage{amssymb}
\usepackage{booktabs}
\title{Amodal Segmentation Based on Visible Region Segmentation and Shape Prior}
\title{Amodal Segmentation Based on Visible Region Segmentation and Shape Prior}
\author {
        Yuting Xiao\textsuperscript{\rm 1, \rm 2},
        Yanyu Xu \textsuperscript{\rm 4},
        Ziming Zhong \textsuperscript{\rm 1, \rm 2},
        Weixin Luo \textsuperscript{\rm 1, \rm 2},
        Jiawei Li \textsuperscript{\rm 3},
        Shenghua Gao \textsuperscript{\rm 1, \rm 2}\\
}
\affiliations {
    \textsuperscript{\rm 1} ShanghaiTech University
    \textsuperscript{\rm 2} Shanghai Engineering Research Center of Intelligent Vision and Imaging
    \textsuperscript{\rm 3} Alibaba Group \\
    \textsuperscript{\rm 4} Institute of High Performance Computing,  A*STAR \\
    \{xiaoyt, zhongzm, luowx, gaoshh\}@shanghaitech.edu.cn, xu\_yanyu@ihpc.a-star.edu.sg, mingong.ljw@alibaba-inc.com
}
\begin{document}

\maketitle

\begin{abstract}
Almost all existing amodal segmentation methods make the inferences of occluded regions by using features corresponding to the whole image.
This is against the human's amodal perception, where human uses the visible part and the shape prior knowledge of the target to infer the occluded region.
To mimic the behavior of human and solve the ambiguity in the learning, we propose a framework, it firstly estimates a coarse visible mask and a coarse amodal mask. Then based on the coarse prediction, our model infers the amodal mask by concentrating on the visible region and utilizing the shape prior in the memory. In this way, features corresponding to background and occlusion can be suppressed for amodal mask estimation. Consequently, the amodal mask would not be affected by what the occlusion is given the same visible regions. The leverage of shape prior makes the amodal mask estimation more robust and reasonable.
Our proposed model is evaluated on three datasets. Experiments show that our proposed model outperforms existing state-of-the-art methods.
The visualization of shape prior indicates that the category-specific feature in the codebook has certain interpretability. The code is available at https://github.com/YutingXiao/Amodal-Segmentation-Based-on-Visible-Region-Segmentation-and-Shape-Prior
\end{abstract}

\section{Introduction}
Amodal segmentation aims to infer the amodal mask, including both the visible region and the possible invisible region of the target object. 
Different from semantic segmentation or traditional instance segmentation, amodal segmentation 
is designed to exploit the amodal perception capability \cite{zhu2017semantic}, where human could infer and perceive the whole semantic concept of the target object mainly according to the partially visible region of the target object. 

Simulating the amodal perception is quite challenging
Lots of efforts have been done for the amodal segmentation, which could be roughly divided into two groups. Methods in the first group directly estimate both the visible and the amodal regions from the images \cite{qi2019amodal, follmann2019learning, zhu2017semantic}, while methods in the second group use inferred depth order information to help the amodal mask prediction \cite{zhang2019learning}.
However, all of them learn the mapping relationship from the feature corresponding to the whole view to the amodal mask.
This processing brings explicitly ambiguity that the same image appearances of occlusion may require different predictions.


\begin{figure*}
    \centering
    \includegraphics[width=0.9\textwidth, height=4cm]{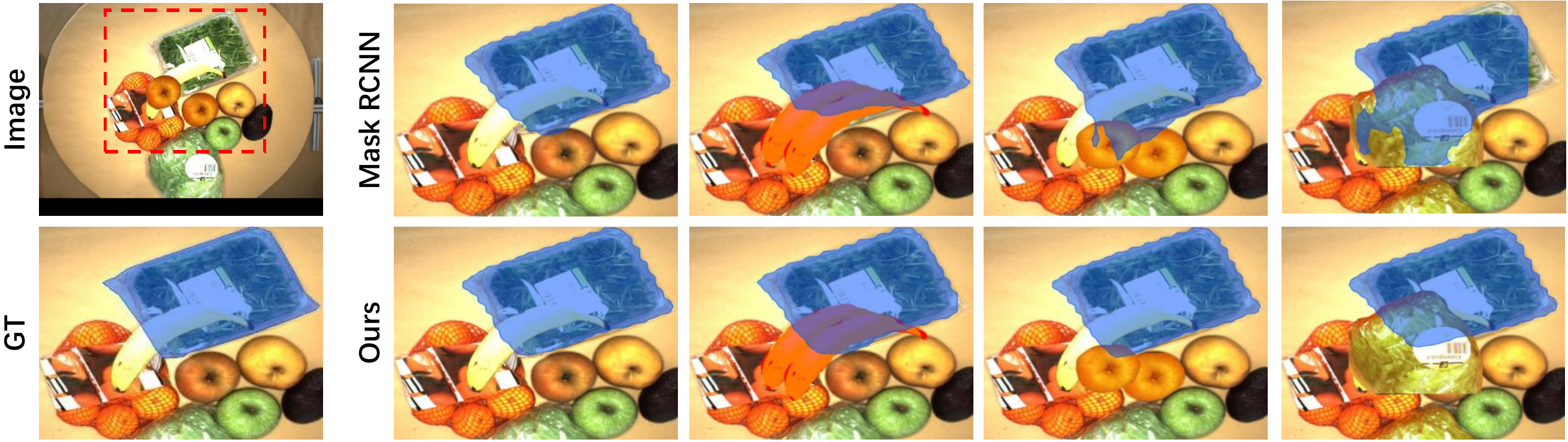}
    \caption{Performance comparison between ours and Mask R-CNN. 
    We overlay different objects on top of the target green vegetables with Adobe Photoshop. Our method could estimate the almost same invisible regions, given different occlusions. For the Mask R-CNN using features corresponding to both occlusions and visible regions, the results are different. 
    }
    \label{fig:teaser}
\end{figure*}

Motivated by the behavior of human leveraging only visible regions and memorizing the category-specific shape prior for amodal segmentation, we propose a solution that estimates the visible region of the target object and leverages the visible region and shape prior for amodal segmentation. 
Without the shape prior of an object, the amodal mask inferred by human might be in an arbitrary shape. Similarly, the estimated amodal mask might also suffer from the lack of shape prior, resulting in the arbitrary edges.
Thus, our model leads to more robust results.
As shown in Fig. \ref{fig:teaser}, although the target (green vegetable) is occluded by different objects, our method could perceive the almost same occluded regions, like human ignoring the different occlusion contexts such as banana, apples, or cabbage. However, the existing baseline estimates the different occluded regions.

In particular, our proposed method consists of a coarse mask segmentation module, a visible mask segmentation module, and an amodal mask segmentation module.
In the coarse mask segmentation module, we utilize the backbone of Mask R-CNN \cite{he2017mask} with an amodal mask head and a visible mask head to predict the coarse amodal and visible mask respectively.
In the visible mask segmentation module, we propose to leverage the amodal mask to refine the visible mask and a reclassification regularizer to alleviate the misleading effect of occlusion in classification. Specifically, we use the coarse amodal mask as the attention multiplying with the feature of region-of-interest for more accurate visible mask estimation. The coarse amodal mask alleviates the effect of the background and contains more information than the coarse visible mask, which provides a cue for the visible mask.
Further, for the reclassification regularizer, we propose to apply the feature of the visible region instead of the whole view to for classification, which alleviates the influence of the occlusion and background.
In the amodal mask segmentation module, we propose to use the feature of the visible region and the shape prior to refine the amodal mask.
Unlike the diversity of the visible mask, the amodal mask has the inherent category-specific shape prior.
To encode the shape prior, we design an auto-encoder for amodal mask encoding, and a codebook is used as the memory obtained from the K-Means of amodal ground-truth mask embeddings. We utilize the shape prior for amodal mask prediction in two aspects: Firstly, we utilize the shape prior to refine the amodal mask. 
Secondly, in the inference, after the final amodal mask prediction, we use the shape prior to post-process the score of proposed boxes to filter out the proposals with a low-quality amodal mask.

As shown in Fig. \ref{fig:teaser}, although the target (green vegetable) is occluded by different objects, our method could perceive the almost same occluded regions with different occlusion contexts such as banana, apples, or cabbage. However, the existing baseline estimates the different occluded regions with features corresponding to different occlusions.

The contributions of this paper could be summarized as followings: (1) inspired by the behavior of human's amodal perception, we propose a novel amodal segmentation model; 
(2) a cross-task attention based refinement strategy is proposed, where the amodal mask and the visible mask is used as attention to refine each other for improving the performance of amodal segmentation; (3) this is the first work that proposes to utilize the shape prior knowledge in the amodal segmentation, and two ways to use shape prior are discussed; (4) experimental results on three public datasets show our method outperforms the existing state-of-the-art methods.

\begin{figure*}[h]
  \centering
  \includegraphics[width=0.9\textwidth, height=6.1cm]{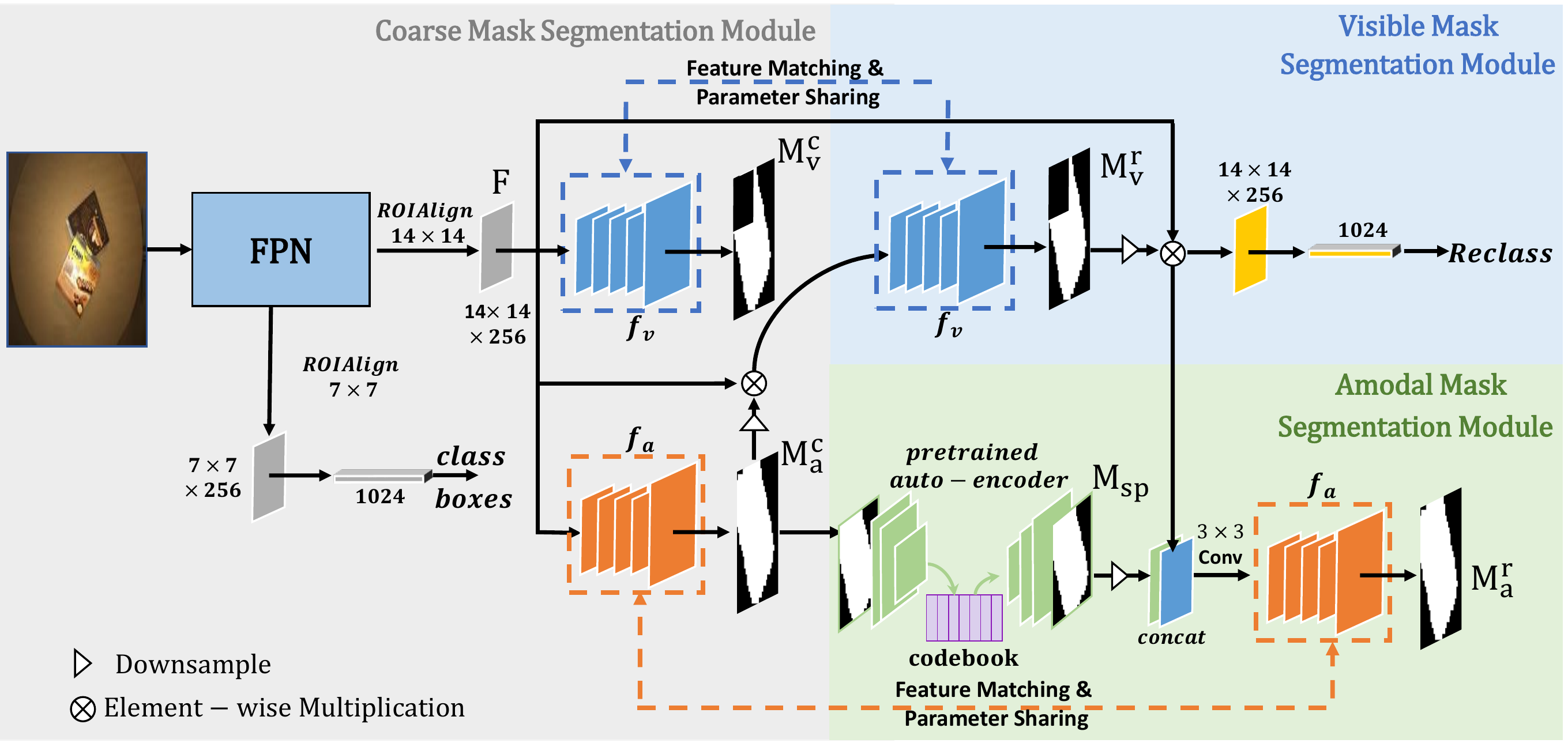}
  \caption{The overview of our approach.}
  \label{fig:overview}
\end{figure*}

\section{Related Work}

\subsection{Visual Occlusion Learning}
Occlusion is inevitable in a large amount of visual task regardless of detection \cite{huang2020nms,xuangeng2020one}, segmentation \cite{fu2019stacked,huang2019ccnet} or inpainting \cite{ren2019structureflow,yu2019free,kar2015amodal} tasks. 
Some researches about occlusion propose interesting and innovative methods or framework to solve the occlusion problem. 
In \cite{yang2019embodied}, Yang \emph{et al.} propose a strategy to see the information behind the occlusion by moving the position of the camera. 
The BANet\cite{chen2020banet} computes the similarity of the pixels in the boundary region of instance to recognize whether these pixels belong to the occlusion or not. 
Some other works tend to explicitly remove the influence of occlusion, the conditional random fields (CRFs) is applied in \cite{John2006layout} to represent the occlusion probability of object parts.
The recent work \cite{huang2020nms,xuangeng2020one} deals with the occlusion problem in human crowd detection which is another task with heavy occlusion. 
The visible region is used to guide the full region of human in \cite{huang2020nms} which indicates that a pair of boxes is generated by a single proposal. 
Besides, in \cite{xuangeng2020one}, Xuan \emph{et al.} applied a similar measure which takes advantage of a proposal corresponding to multiple predictions to avoid the ambiguous regression.
In 3D area, many methods \cite{zhang2020object}, \cite{ramamonjisoa2020predicting,cheng2019occlusion} about solving the negative influence of occlusion are utilized for 3D keypoint, mesh and depth estimation.

\subsection{Amodal Segmentation}
A large number of researches has been accomplished on instance segmentation \cite{bai2017deep,ren2017end,xu2019explicit,neven2019instance,chen2019hybrid}, based on classic detection framework such as Faster-RCNN \cite{ren2015faster} or YOLO \cite{redmon2016you}. 
Mask R-CNN \cite{he2017mask}, one of the most representative two-stage approach, uses a mask head added in the Faster-RCNN to process the aggregated feature sampled by the ROIAlign module. 
In order to utilize the multiple scale information, the FPN \cite{lin2017feature} is proposed to detect instances on different scales. 
Further, the Path aggregation network \cite{liu2018path} is proposed to boost information flow in feature hierarchy. 
Besides, some other works localize the position of instances by the center point of each instance \cite{xie2019polarmask} instead of the boundary box or iteratively deform an initial contour to the boundary box \cite{peng2020deep}. 

As a newly developing direction of instance segmentation,
The earliest work about amodal segmentation is proposed by \cite{li2016amodal}, which utilizing generated data by overlapping instances on other instances to train and test their method. They extend the boxes of each instance and refine the heatmap. 
And in \cite{zhu2017semantic}, the SharpMask \cite{pinheiro2015learning} which predicts the object mask from coarse to fine is provided as the baseline model. 
Recently, several amodal segmentation datasets are released to help for the amodal segmentation research. 
The ORCNN \cite{follmann2019learning} uses a visible mask head and an amodal mask head to directly predict the visible mask and amodal mask respectively, obtaining the occlusion mask by subtracts the visible mask from the amodal mask.
The SLN \cite{zhang2019learning} modeling a depth order representation to infer the amodal mask.
The method proposed in \cite{qi2019amodal} utilizes an occlusion classifier to recognize whether an instance is occluded and ensemble the feature from box and class head by multi-level coding.


\section{Method}

\subsection{Problem Formulation}

Given an image $\mathbf{I}$, amodal segmentation aims to estimate the amodal mask $\mathbf{M}_a$ as well as the visible mask $\mathbf{M}_v$ for a region-of-interest (ROI). 
The visible mask $\mathbf{M}_v$ could be estimated directly from the image. The amodal mask $\mathbf{M}_a$ consists of both the visible part and the invisible part. The most challenging part of amodal segmentation is to estimate the invisible region based on the visible region and without being affected by the occlusions.  

The human amodal perception means the ability to infer the global instance according to the partial observation.
When inferring the invisible region, human utilizes the feature of the visible region and the shape prior of the object.
Inspired by this, we formulate the amodal segmentation as learning a nonlinear mapping function that maps the ROI feature $\mathbf{F}$ of an image $\mathbf{I}$ to the visible mask $\mathbf{M}_v$ and the amodal mask $\mathbf{M}_a$ with the regularization of shape prior.
As shown in Fig. \ref{fig:overview}, our proposed method consists of a coarse mask segmentation module, a visible mask segmentation module, and an amodal mask segmentation module. 

\subsection{The Coarse Mask Segmentation Module}

The coarse mask segmentation module aims to extract the target visual features and predict the coarse amodal mask $\mathbf{M}_a^c$ and the coarse visible mask $\mathbf{M}_v^c$. 
Following the existing works \cite{zhang2019learning, follmann2019learning, qi2019amodal}, we also employ a ResNet50 \cite{he2016deep} based FPN \cite{lin2017feature} as the backbone to extract the visual feature of the region-of-interest $\mathbf{F}$ containing the target object.
Specifically, in this module, both the visible mask head $f_v$ and the amodal mask head $f_a$ take the feature $\mathbf{F}$ as input.
The amodal mask head and the visible mask head have the same network structure that consists of 4 convolution layers and 1 deconvolution layer with different parameters.

There are four loss terms in this module, including a coarse amodal mask loss $\mathcal{L}_{BCE}(\mathbf{M}_a^c, \mathbf{M}_a^g)$, a coarse visible mask loss $\mathcal{L}_{BCE}(\mathbf{M}_v^c, \mathbf{M}_v^g)$, a classification loss $\mathcal{L}_{\text{cls}}$ and an object bounding box regression loss $\mathcal{L}_{\text{reg}}$.The $\mathbf{M}_a^g$ and $\mathbf{M}_v^g$ are the amodal and visible ground-truth mask. Both of the $\mathcal{L}_{cls}$ and $\mathcal{L}_{reg}$ are the same as the loss function of the class and box head in Mask R-CNN \cite{he2017mask}. The $\mathcal{L}_{BCE}(\cdot, \cdot)$ is the binary cross-entropy loss.

\subsection{The Visible Mask Segmentation Module}

The visible segmentation module aims to further refine the visible mask, via the amodal mask and the reclassification regularizer. 
Because the amodal mask contains the visible region, we use it as the attention to multiple with the ROI feature $\mathbf{F}$ for the visible mask refinement. This operation enhances the capability to distinguish occlusion and target instance of the visible mask head.  Further, it also alleviates the effect of background features for visible mask refinement.
The loss term of visible mask refinement is
\begin{equation}
\begin{aligned}
 \mathcal{L}_v^r & = \frac{1}{N}\begin{matrix} \sum_i^N \end{matrix}
        \mathcal{L}_{BCE}(f_v(\mathbf{F}_i \cdot \mathbf{M}_{a,i}^c), \mathbf{M}_{v,i}^g),
\end{aligned}
\end{equation}
where $N$ is the number of predicted instances. We denote the refined visible mask of $i^{th}$ instance as $\mathbf{M}^r_{v,i}$, which is the output of the visible mask head $f_v$ in this term.

The reclassification regularizer aims to classify each instance by processing the feature of the visible region, which avoids the misleading effect of the feature corresponding to the occlusion and background. The input of the reclassification regularizer is the feature of refined visible region regarded as $\mathbf{F}\cdot \mathbf{M}_v^r$. The reclassification regularizer $f_{rc}$ composes of two fully connected layers. In the inference, we obtain the class score of each instance by multiplying the score of the class head and reclassification regularizer.
The loss term of the reclassification regularizer is
\begin{equation}
\begin{aligned}
\mathcal{L}_{rc} =  \frac{\lambda_{rc}}{N}\begin{matrix} \sum_i^N \end{matrix}\mathcal{L}_{CE}(f_{rc}(\mathbf{F}_i \cdot \mathbf{M}_{v,i}^r), y_i),
\end{aligned}
\end{equation}
where $y_i$ is the class label of $i^{th}$ instance. $\mathcal{L}_{CE}(\cdot,\cdot)$ is the cross-entropy loss. The the hyper-parameter $\lambda_{rc}=0.25$.

\subsubsection{Feature matching}
Feature matching is commonly used in network model acceleration and compression \cite{Ba2013Do, Li_2017_CVPR}, where a compact student model mimics the feature maps extracted from a large teacher model to improve its accuracy. 
In this work, we also employ the feature matching strategy for the visible mask head to reduce the gaps between feature maps extracted in the coarse mask prediction and the refined mask prediction.  
Since the visible mask head has the same network structure and the same parameters in predicting the coarse and refined mask, and the only difference is the inputs. 
The feature matching loss between the coarse visible mask prediction and refined visible mask prediction helps the network concentrate more on visible region appearance for visible mask segmentation and alleviate the effect of the background. 

In our implementation, we use feature in the last two convolution layers of the visible mask head to measure the feature matching loss. We denote the loss function of the feature matching of the visible mask head as $ L_{vfm}$, and use a subscript ($j$) on the visible mask head $f_v$ to denote the feature maps in the $i^{\text{th}}$ layer. Here $j=4,5$, which means we only use features in the higher level convolution layers. Then the feature matching loss of the visible mask head is
\begin{equation}
\begin{aligned}
        \mathcal{L}_{vfm} = \frac{1}{N\cdot S}\begin{matrix}\sum_{i,j}^{N,S}\end{matrix}
        \lambda_j\mathcal{L}_S(f^{(j)}_v(\mathbf{F}_i),f^{(j)}_v(\mathbf{F}_i\cdot \mathbf{M}_{a,i}^c)),
\end{aligned}
\end{equation}
where $N,S$ is the number of instances and the number of convolution layers of the visible mask head respectively.
And we apply cosine similarity $\mathcal{L}_S$ \cite{zhu2019deformable} in feature matching. We set the hyper-parameters $\lambda_4=0.01$, $\lambda_5=0.05$ and $\lambda_j=0 (j\in\{1,2,3\})$. 

\subsection{The Amodal Mask Segmentation Module}
The amodal mask segmentation module is designed to refine the coarse amodal mask by using the feature of the visible region and the shape prior. Inferring the amodal mask from the visible region appearance helps our model alleviate the misleading effect of the occlusion feature. Besides, different from the visible mask, for each category, the shapes of amodal masks are explicitly more stable, without the effect of arbitrary occlusions.
The feature of the visible region can be obtained by using the refined visible mask $\mathbf{M}_v^r$ from the visible mask segmentation module as visible attention.


\begin{figure}[t]
    \centering
    \includegraphics[height=4cm, width=0.48\textwidth]{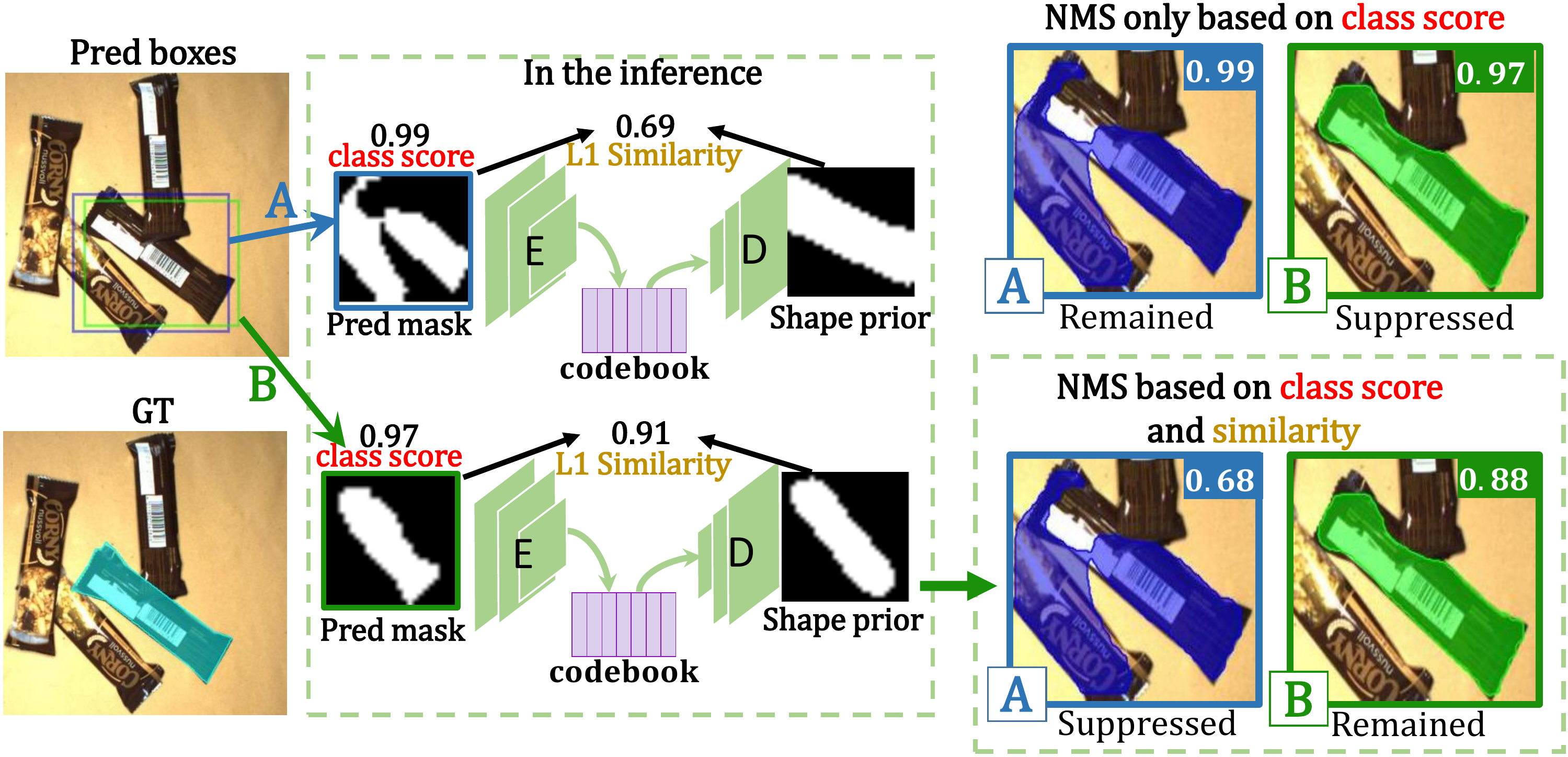}
    \caption{The illumination of the shape prior post-process. The predicted amodal mask of box B gets a higher IoU value than the predicted amodal mask of box A. But the class score of box B (0.97) is lower than box A (0.99), which results in the suppression of box B in NMS. The shape prior post-process multiplies the shape prior similarity with the class score to refine the score of object. The score of A is $0.99\times0.69 \approx 0.68$ while the score of B is $0.97\times 0.91 \approx 0.88$. The box B is remained while the box A is suppressed by NMS.}
    \label{fig:shape prior}
\end{figure}

In the pre-training phase, to model the shape prior, we collect the amodal ground-truth masks in the training set and use an auto-encoder to obtain the embedding of each amodal mask. Then, for each category, we partition the embeddings into $K$ clusters via K-Means clustering and use the centers of these clusters as a codebook. Thus the codebook could memorize the embeddings of category-specific shape prior. This codebook is used to further refine the amodal mask. 

\begin{table*}[h]
\centering
\resizebox{\textwidth}{14mm}{
\begin{tabular}{l|ccccc|cc|ccccc|cc|ccccc|cc}
\toprule
\multirow{3}{*}{Methods} & \multicolumn{7}{c|}{D2SA}                                                                                                                                     & \multicolumn{7}{c|}{KINS}                                                                                                                                     & \multicolumn{7}{c}{COCOA cls}                                                                                                                                \\ \cline{2-22} 
                         & \multicolumn{5}{c|}{Amodal}                                                                                                 & \multicolumn{2}{c|}{Visible}    & \multicolumn{5}{c|}{Amodal}                                                                                                 & \multicolumn{2}{c|}{Visible}    & \multicolumn{5}{c|}{Amodal}                                                                                                 & \multicolumn{2}{c}{Visible}    \\ \cline{2-22} 
                         & AP             & AP50           & AP75           & AR             & \begin{tabular}[c]{@{}c@{}}AP\\ (Occluded)\end{tabular} & AP             & AR             & AP             & AP50           & AP75           & AR             & \begin{tabular}[c]{@{}c@{}}AP\\ (Occluded)\end{tabular} & AP             & AR             & AP             & AP50           & AP75           & AR             & \begin{tabular}[c]{@{}c@{}}AP\\ (Occluded)\end{tabular} & AP             & AR             \\ \hline
Mask RCNN                & 63.57          & 83.85          & 68.02          & 65.18          & NA                                                       & 68.98          & 70.11          & 30.01          & 54.53          & 30.11          & 19.42          & NA                                                       & 28.00          & 19.23          & 33.67          & 56.50          & 35.78          & 34.18          & NA                                                       & 30.10          & 31.52          \\
Mask RCNN(C8)            & 64.85          & 84.05          & 70.72          & 65.61          & NA                                                       & 69.81          & 70.54          & 30.71          & 54.36          & 31.47          & 19.77          & NA                                                       & 28.74          & 19.38          & 34.72          & \textbf{57.50} & 36.93          & 35.45          & NA                                                       & 31.89          & 32.88          \\
ORCNN                    & 64.22          & 83.55          & 69.12          & 65.25          & 45.27                                                   & 69.67          & 70.46          & 30.64          & 54.21          & 31.29          & 19.66          & 34.23                                                   & 28.77          & 20.01          & 28.03          & 53.68          & 25.36          & 29.83          & 17.40                                                   & 30.80          & 32.23          \\
SLN                      & 25.10          & 30.80          & 29.40          & 19.20          & NA                                                       & NA              & NA              & 6.60          & 10.70         & 6.90          & 6.10            & NA                                                       & NA              & NA              & 14.40          & 23.60          & 15.80          & 17.10          & NA                                                       & NA              & NA              \\
Our method               & \textbf{70.27} & \textbf{85.11} & \textbf{75.81} & \textbf{69.17} & \textbf{51.17}                                          & \textbf{72.28} & \textbf{71.85} & \textbf{32.08} & \textbf{55.37} & \textbf{33.34} & \textbf{20.90} & \textbf{37.40}                                          & \textbf{29.88} & \textbf{19.88} & \textbf{35.41} & 56.03          & \textbf{38.67} & \textbf{37.11} & \textbf{22.17}                                          & \textbf{34.58} & \textbf{36.42} \\ 
\bottomrule
\end{tabular}
}
\caption{The comparison on the D2SA dataset, the KINS dataset, and the COCOA cls dataset. Because some methods only output the amodal mask prediction, the AP (Occluded) and visible mask prediction performance of them are unavailable (NA).}
\label{tab:exp_comparison}
\end{table*}

In the training phase, for a predicted coarse amodal mask $\mathbf{M}_a^c$, we feed it into the encoder of the pre-trained auto-encoder to obtain the embedding.
We use an L2 distance to find  $k$ nearest embeddings in the category-specific codebook according to the predicted category. This is denoted as shape prior search to obtain the shape prior embeddings.
Then we feed these $k$ nearest embeddings into the decoder to get decoded shape prior masks $\mathbf{M}_{sp}^k=f_{sp}(\mathbf{M}_a^c)$ ($\mathbf{M}_a^c$$\in$$\mathbb{R}^{H\times W}$, $\mathbf{M}_{sp}^k$$\in$$\mathbb{R}^{k\times H\times W}$). The $f_{sp}$ denotes the operation using an auto-encoder with category-specific codebook for shape prior search.
From this operation, we obtain the category-specific shape prior masks $\mathbf{M}_{sp}^k$ which are the most similar to the coarse amodal mask in the shape prior.
Then, we concatenate the feature of the visible region $\mathbf{F}\cdot\mathbf{M}_v^r$ and the $k$ nearest shape prior amodal masks $\mathbf{M}_{sp}^k$ as the input of the amodal mask head for amodal mask refinement. This process is to imitate the perception of human that infers the amodal mask objects by focusing on the appearance at the visible region and using the shape prior knowledge. The total loss term of this processing can be denoted by
\begin{equation}
\begin{aligned}
    \mathcal{L}_a^r =  \frac{1}{N} \begin{matrix}\sum_i^N\end{matrix}
    \mathcal{L}_{CE}(f_a(cat(\mathbf{F}_i\cdot \mathbf{M}_{v,i}^r, \mathbf{M}_{sp,i}^k)), \mathbf{M}_{a,i}^g),
\end{aligned}
\end{equation}
where the $f_a$ is the amodal mask head whose output is $\mathbf{M}_a^r$. The $cat(\cdot,\cdot)$ is the matrix concatenate operation. In our implementation, we set $k=16$.
Besides, similar to the equation (3) of the visible segmentation module, we also apply feature matching to the amodal mask head to enhance the capacity of focusing on the visible region. The loss function of feature matching in the amodal mask head is
\begin{equation}
\begin{aligned}
    \mathcal{L}_{afm} = \frac{1}{N\cdot S} \begin{matrix}\sum_{i,j}^{N,S}\end{matrix}\lambda_j\mathcal{L}_S(f^{(j)}_a(\mathbf{F}_i),f^{(j)}_a(\mathbf{F}_i\cdot \mathbf{M}_{v,i}^r)).
\end{aligned}
\end{equation}

In the inference phase, the shape prior can also be used to help the network further improve amodal segmentation.
As shown in Fig. \ref{fig:shape prior}, we use the difference between the refined amodal mask and its nearest counterpart in the codebook $\|\mathbf{M}_a^r - \mathbf{M}_{sp}^k\|_2$ as a measurement to rank the scores of bounding box proposals, and filter out the proposals with larger differences. Here the difference is measured by an L1 distance. Here we set $k=1$.
If the predicted amodal mask shape of an instance is explicitly different from the nearest shape prior in the memory, it should be treated as a low-quality prediction even with a high class score.

 \begin{figure*}
  \centering
  \includegraphics[width=0.85\textwidth, height=8cm]{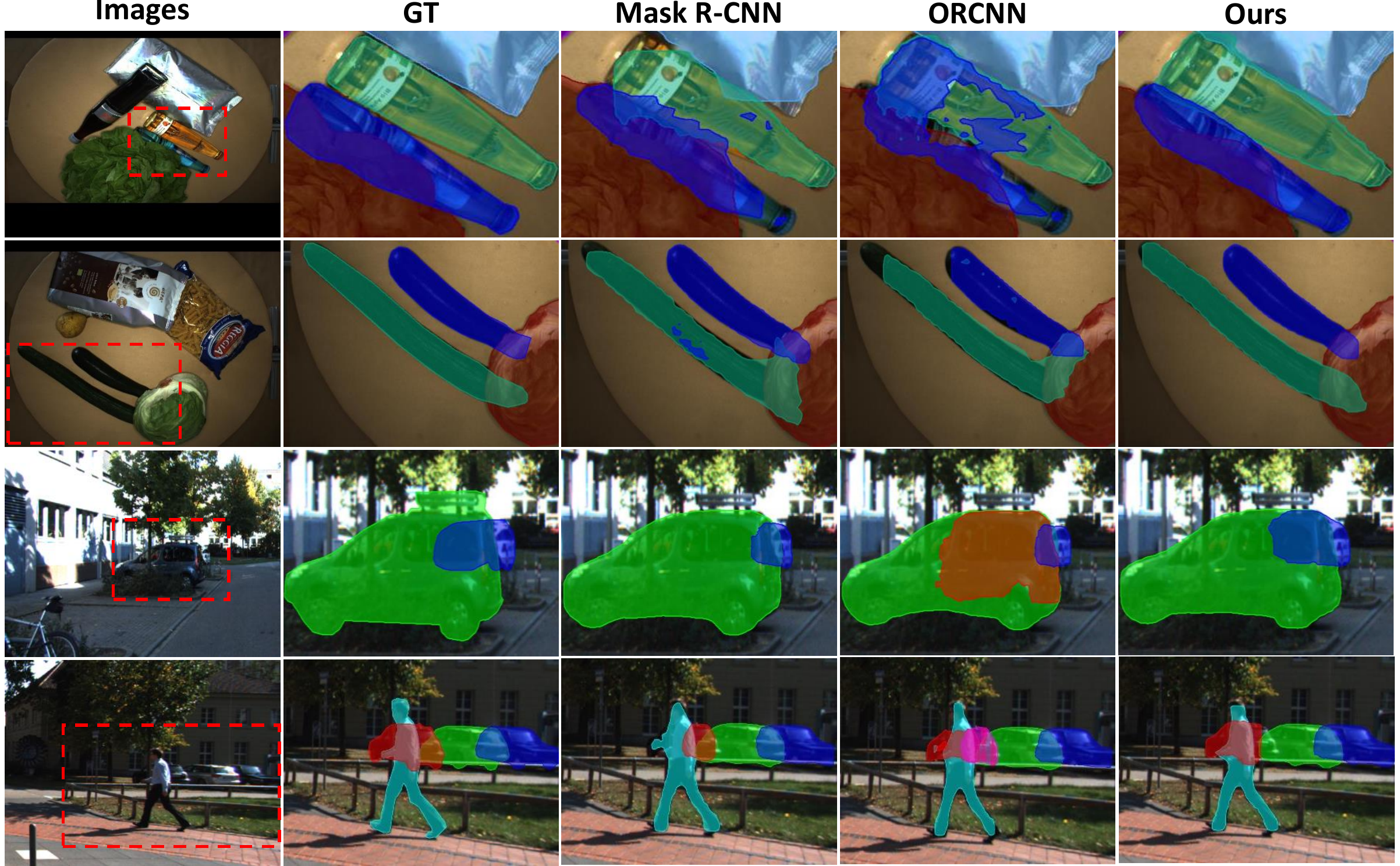}
  \caption{The columns from left to right are the images, the ground-truth amodal masks, estimations of Mask R-CNN, ORCNN and ours, respectively.}
  \label{fig:exp_examples}
\end{figure*}

\subsection{The Implementation Details}
Our model could predict the amodal mask and the visible mask.
In the visible mask segmentation module and amodal mask segmentation module, we use the coarse amodal mask and refined visible mask as attention. However, if directly using the predicted masks in the warm-up phase, these inaccurate predictions might destroy the following parts. Thus, we design a weighting operation, where each instance is assigned an amodal weight and a visible weight to measure the weight of an instance in optimization. 
This operation can be found in supplementary. The final loss function is
\begin{equation}
    \mathcal{L} = \mathcal{L}_{cls} + \mathcal{L}_{reg} + \mathcal{L}_a^c + \mathcal{L}_v^c + \mathcal{L}_a^r + \mathcal{L}_v^r + \mathcal{L}_{rc} + \mathcal{L}_{afm} + \mathcal{L}_{vfm}.
\end{equation}
Stochastic Gradient Descent (SGD) \cite{zinkevich2010parallelized} with weight decay is used for optimization in the training.

\begin{table*}[h]
\centering
\resizebox{0.93\textwidth}{14mm}{
\begin{tabular}{lccccl|ccc|cc|ccc|cc}
\toprule
\multirow{2}{*}{} & \multicolumn{5}{c|}{\multirow{2}{*}{Ablation Study}}                                                                                                                                                                                                                                                & \multicolumn{5}{c|}{D2SA}                                                                                                                                                                               & \multicolumn{5}{c}{KINS}                                                                                                   \\ \cline{7-16} 
                  & \multicolumn{5}{c|}{}                                                                                                                                                                                                                                                                               & \multicolumn{3}{c|}{Amodal}                                                               & \multicolumn{2}{c|}{Visible}                                                                                & \multicolumn{3}{c|}{Amodal}                                                               & \multicolumn{2}{c}{Visible}    \\ \hline
                  & \begin{tabular}[c]{@{}c@{}}Visible\\ Attention\end{tabular} & Reclass    & \begin{tabular}[c]{@{}c@{}}Shape Prior\\ Refinement\end{tabular} & \begin{tabular}[c]{@{}c@{}}Shape Prior\\ Post-process\end{tabular} & \multicolumn{1}{c|}{\begin{tabular}[c]{@{}c@{}}Feature\\ Matching\end{tabular}} & AP             & AR             & \begin{tabular}[c]{@{}c@{}}AP\\ (Occluded)\end{tabular} & \begin{tabular}[c]{@{}c@{}}Visible\\ AP\end{tabular} & \begin{tabular}[c]{@{}c@{}}Visible\\ AR\end{tabular} & AP             & AR             & \begin{tabular}[c]{@{}c@{}}AP\\ (Occluded)\end{tabular} & AP             & AR             \\ \hline
\textbf{1}        &                                                             & \checkmark & \checkmark                                                       & \checkmark                                                         &                                                                                 & 66.67          & 65.68          & 46.40                                                   & 70.50                                                & 70.24                                                & 31.61          & 20.16          & 36.84                                                   & 29.44          & 19.61          \\
\textbf{2}        & \checkmark                                                  &            & \checkmark                                                       & \checkmark                                                         & \multicolumn{1}{c|}{}                                                           & 69.02          & 68.09          & 50.31                                                   & 71.66                                                & 70.83                                                & 31.65          & 20.55          & 36.93                                                   & 29.25          & 19.81          \\
\textbf{3}        & \checkmark                                                  & \checkmark &                                                                  & \checkmark                                                         &                                                                                 & 68.84          & 67.42          & 49.08                                                   & 71.53                                                & 70.70                                                & 31.70          & 20.33          & 37.25                                                   & 29.45          & 19.63          \\
\textbf{4}        & \checkmark                                                  & \checkmark & \checkmark                                                       &                                                                    &                                                                                 & 68.15          & 68.30          & 50.03                                                   & 70.71                                                & 70.43                                                & 31.87          & 20.46          & 37.48                                                   & 29.53          & 19.70          \\
\textbf{5}        & \checkmark                                                  & \checkmark & \checkmark                                                       & \checkmark                                                         &                                                                                 & 69.98          & 68.87          & 51.01                                                   & 71.92                                                & 71.15                                                & 31.94          & 20.60          & 37.55                                                   & 29.61          & \textbf{19.88} \\
\textbf{6}        & \checkmark                                                  & \checkmark & \checkmark                                                       & \checkmark                                                         & \multicolumn{1}{c|}{\checkmark}                                                 & \textbf{70.27} & \textbf{69.17} & \textbf{51.17}                                          & \textbf{72.28}                                       & \textbf{71.85}                                       & \textbf{32.08} & \textbf{20.90} & \textbf{37.57}                                          & \textbf{29.88} & \textbf{19.88} \\ 
\bottomrule
\end{tabular}
}

\caption{The ablation studies results on the D2SA dataset and the KINS dataset.}
\label{tab:exp_ablation}
\end{table*}
\section{Experiments}

\subsection{Experimental Setting}
We implement our proposed model based on Detectron2 \cite{wu2019detectron2} on the PyTorch framework.
The main parameter setting is: For the D2SA dataset, batch size(2), learning rate (0.005), and the number of iteration (70000). For the KINS dataset, batch size(1), learning rate (0.0025), and the number of iteration (48000). For the COCOA cls dataset, batch size (2), learning rate (0.0005), and the number of iteration (10000).

\textbf{Datasets.} We evaluate the model performance for amodal segmentation on three
datasets: the D2SA (D2S amodal) \cite{follmann2019learning}, the KINS dataset \cite{qi2019amodal}, the COCOA cls dataset \cite{zhu2017semantic}.

The D2SA dataset is built based on the D2S(Densely Segmented Supermarket) dataset with 60 categories of instances. It contains 2000 images in the training set and 3600 images in the validation set, where the annotations of the amodal mask are generated by overlapping one to another.

The KINS dataset is built based on the KITTI dataset \cite{geiger2012we}. It consists of 7474 images in the training set and 7517 images in the validation set. Different from the D2SA dataset, its amodal ground-truth is manually annotated. 
There are 7 categories about the autonomous driving task in the KINS dataset.

The COCOA cls dataset \cite{zhu2017semantic} is built based on the COCO dataset \cite{lin2014microsoft}. It consists of 2476 images in the training set and 1223 images in the validation set. There are 80 categories in this dataset.

\textbf{Metrics.} 
Following the \cite{zhu2017semantic, zhang2019learning}, we use the mean average precision (AP) and mean average recall (AR) to evaluate performances. The AP (Occluded) is computed on the instances whose occlusion rate is larger than 15\%.
We use the evaluation api of the COCO dataset \cite{lin2014microsoft} for fair comparisons.

\textbf{Baselines.}
We use following state-of-the-art methods for comparison.

(1) Mask R-CNN\cite{he2017mask} predicts the amodal masks by a mask head consisting of 4 convolution layers and 1 deconvolution layer. We train two Mask R-CNN baselines predicting amodal mask and visible mask respectively since the Mask R-CNN cannot predict both of them simultaneously. The Mask R-CNN (C8) means that the mask head has 8 convolution layers. 


(2) ORCNN \cite{follmann2019learning} uses an amodal mask head and a visible mask head to infer the amodal masks and visible maks. The invisible mask is obtained by abstracting the visible mask from the amodal mask. 

(3) SLN \cite{zhang2019learning} claims the importance of depth information in amodal segmentation. It uses a semantics-aware distance map to predict the amodal mask by utilizing depth order information.

\subsection{Performance Comparison}
We compare our model with all comparative methods on the datasets mentioned, and the performance comparisons are shown on Table \ref{tab:exp_comparison}.
We can see that our model always outperforms other methods. Compared with Mask R-CNN and Mask R-CNN (C8), the improvement resulting from directly adding the depth of the mask head is explicitly lower than the improvement achieved by our model. Our method gets better performance mainly due to our reasonable design of the network rather than the expansion of the network. 

We also show some qualitative results estimated by Mask R-CNN, ORCNN, and our method in Fig .\ref{fig:exp_examples}. 
We can see that our method can segment more accurately than other methods, owning to the help of the attention on the visible region and the shape prior. For the 1st and 2nd row, the predictions of our method are not misled by the feature of occlusions such as the glass bottle and cabbage. For the 3rd and 4th row, our method keeps robust even the occlusion rate is large.

\subsection{Ablation Studies}

We conduct the ablation studies on both the D2SA dataset and the KINS dataset. All the results are shown on Table \ref{tab:exp_ablation}.

\subsubsection{The Effect of Visible Attention.}
To evaluate the effect of visible attention in refining the amodal mask, we design the baseline refining the amodal mask without using the visible mask as attention. The input of the amodal mask head for predicting refined amodal mask is the concatenation of ROI feature $\mathbf{F}$ and shape prior $\mathbf{M}_{sp}^k$. Experimental results are shown at the 1st and 5th rows on Table \ref{tab:exp_ablation}.

\begin{table*}[h!]
\centering
\resizebox{0.9\textwidth}{12mm}{
\begin{tabular}{l|ccccc|cc|ccccc|cc}
\toprule
\multirow{3}{*}{Types of Utilizing Attention} & \multicolumn{7}{c|}{D2SA}                                                                                                                                     & \multicolumn{7}{c}{KINS}                                                                                                                                     \\ \cline{2-15} 
                                                & \multicolumn{5}{c|}{Amodal}                                                                                                 & \multicolumn{2}{c|}{Visible}    & \multicolumn{5}{c|}{Amodal}                                                                                                 & \multicolumn{2}{c}{Visible}    \\ \cline{2-15} 
                                                & AP             & AP50           & AP75           & AR             & \begin{tabular}[c]{@{}c@{}}AP\\ (Occluded)\end{tabular} & AP             & AR             & AP             & AP50           & AP75           & AR             & \begin{tabular}[c]{@{}c@{}}AP\\ (Occluded)\end{tabular} & AP             & AR             \\ \hline
(a) Both Self Attention                          & 64.55          & 82.53          & 68.96          & 65.62          & 44.31                                                   & 70.50          & 71.02          & 31.18          & 54.24          & 32.08          & 20.15          & 36.89                                                   & 29.27          & 19.76          \\
(b) Only Visible Attention                       & 66.98          & 83.99          & 72.87          & 67.78          & 47.53                                                   & 70.92          & 71.50          & 31.32          & \textbf{54.79} & 32.23          & 20.25          & 36.98                                                   & 29.40          & 19.81          \\
(c) Cross Attention                              & 67.11          & 83.90          & 72.81          & 67.70          & 47.75                                                   & 71.79          & 71.77          & 31.57          & 54.58          & 32.54          & 20.61          & 37.13                                                   & 29.65          & \textbf{20.09} \\
(d) Ours                                         & \textbf{67.33} & \textbf{84.10} & \textbf{72.97} & \textbf{68.06} & \textbf{47.91}                                          & \textbf{71.88} & \textbf{72.43} & \textbf{31.69} & 54.52          & \textbf{32.96} & \textbf{20.75} & \textbf{37.30}                                          & \textbf{29.68} & 20.01          \\
\bottomrule
\end{tabular}
}
\caption{The experimental results of the attention analysis on the D2SA dataset and the KINS dataset.}
\label{tab:exp_cross-task}
\end{table*}

\subsubsection{The Effect of Reclassification Regularizer.}
To validate the effectiveness of the reclassification regularizer. We conduct the experiments at the 2nd and 5th rows on Table \ref{tab:exp_ablation}. 
The experimental results show the importance of the reclassification regularizer.


\subsubsection{The Effect of the Shape Prior Refinement.}
To investigate the effect of shape prior in refinement, we plan to train the baseline at the 3rd without shape prior refinement. We only utilize the feature of the refined visible region to refine the amodal mask.
The gap between the results at the 3rd row the 5th row shows the importance of shape prior knowledge, which agrees with the usage of shape prior in the human's amodal perception.

\begin{figure*}[t]
\begin{center}
	\begin{tabular}{ccc}
		\includegraphics[height=3.8cm,width=0.28\textwidth]{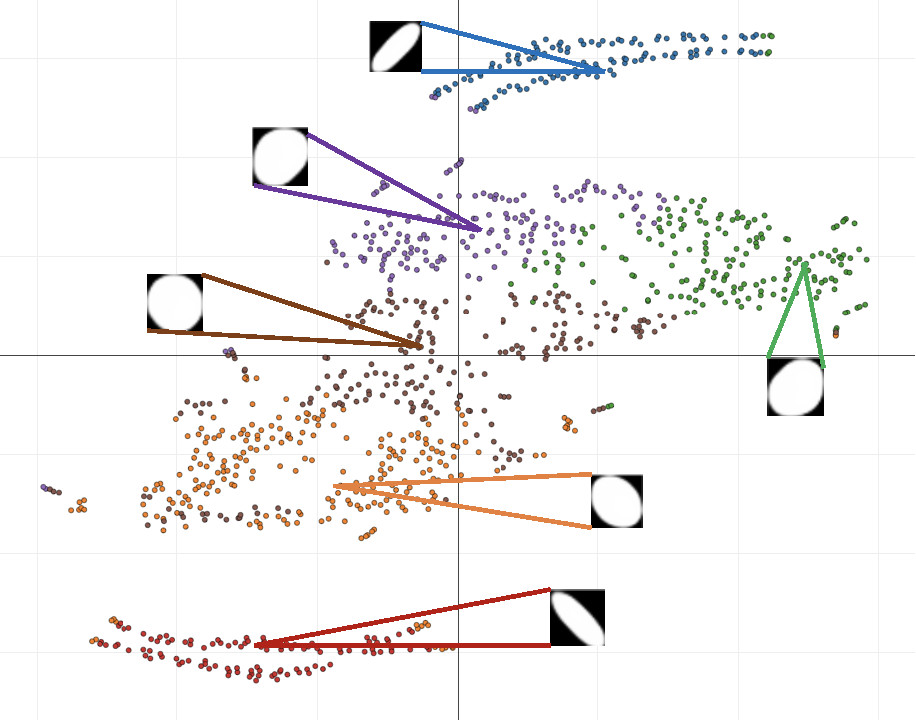} &
		\includegraphics[height=3.8cm,width=0.28\textwidth]{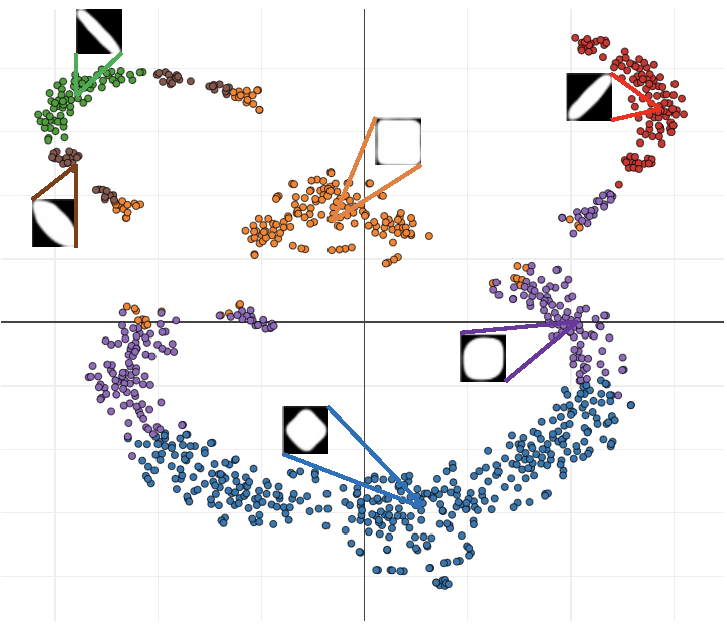} &
		\includegraphics[height=3.8cm,width=0.28\textwidth]{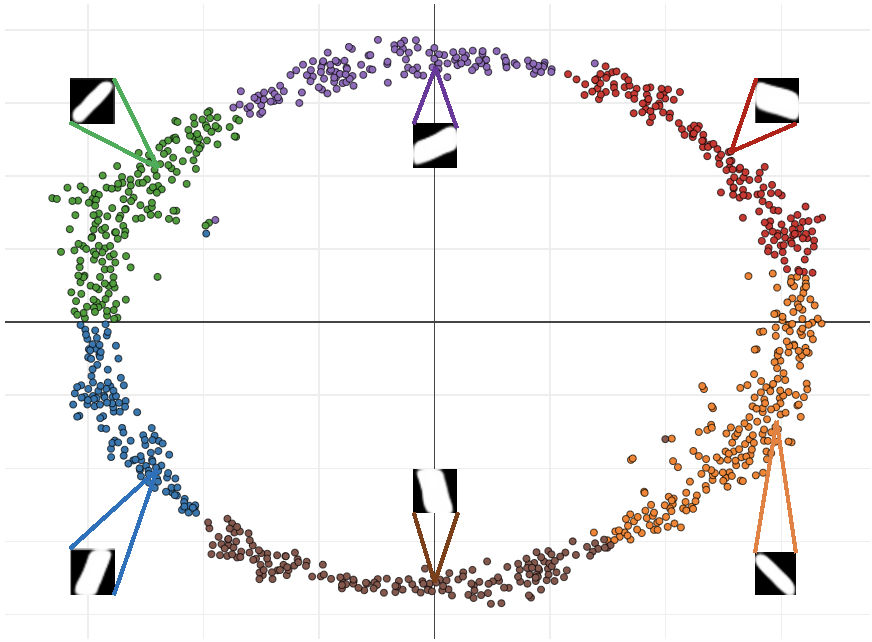}\\
		(a) Carrot  & (b) Bottle & (c) Cucumber \\
	\end{tabular}
\end{center}
\caption{Visualization of shape prior embeddings in shape codebook. We use t-SNE \cite{maaten2008visualizing} to visualize the latent space of each category in the codebook. This proves that the learned shape prior has certain interpretability.}
\label{fig:exp_kmeans}
\end{figure*}

\subsubsection{The Effect of the Shape Prior Post-process.}
To investigate the effect of shape prior post-process, we plan to remove the shape prior post-process based on the baseline at the 5th row. The result is shown at the 4th row. Compared with the results at the 5th row, the gap shows that the utilizing of shape prior post-process can achieve improvement.

\subsubsection{The Effect of Feature Matching.}
To validate the effect of feature matching in our method, we train a baseline via removing the feature matching on both the amodal and visible mask head. The result is shown in the 5th row on Table \ref{tab:exp_ablation}. Compared with our model at 6th row, the feature matching could achieve further better performance, which indicates that the feature matching could further help the amodal mask head learn the feature extracted from the visible regions.

\subsection{Attention Analysis}
To evaluate the effect of amodal attention and visible attention in our model, we propose 4 different types to utilize amodal and visible attention. The details are shown in supplementary. We conduct the experiments as shown on Table \ref{tab:exp_cross-task} to compare different ways to use amodal and visible attention. The shape prior and reclassification regularizer are removed in this section.

\textbf{(a) Both Self-Attention} uses the coarse amodal mask and coarse visible mask as attention to refine amodal mask and visible mask respectively. The input of the amodal mask head and the visible mask head in attention-based refinement are the feature of the amodal region $\mathbf{F}\cdot\mathbf{M}_a^c$ and the feature of the visible region $\mathbf{F}\cdot\mathbf{M}_v^c$ respectively.

\textbf{(b) Only Visible Attention} uses the coarse visible mask as attention to refine the visible mask prediction, which indicates $\mathbf{M}_v^r = f_v(\mathbf{F}\cdot\mathbf{M}_v^c)$. Then, the refined visible mask $\mathbf{M}_v^r$ is utilized as attention to refine amodal mask prediction. The refined amodal mask is obtained by $\mathbf{M}_a^r = f_a(\mathbf{F}\cdot\mathbf{M}_v^r)$.

\textbf{(c) Cross Attention} uses the coarse amodal mask and coarse visible mask as attention to refine the visible and amodal mask prediction respectively. The formula is $\mathbf{M}_a^r=f_a(\mathbf{F}\cdot\mathbf{M}_v^c)$ and $\mathbf{M}_v^r=f_v(\mathbf{F}\cdot\mathbf{M}_a^c)$.

\textbf{(d) Ours} uses the coarse amodal mask as attention to refine the visible mask, $\mathbf{M}_v^r = f_v(\mathbf{F}\cdot\mathbf{M}_a^c)$. Then, using the refined visible mask as attention to refine the amodal mask, $\mathbf{M}_a^r = f_a(\mathbf{F}\cdot\mathbf{M}_v^r)$.

We can observe that utilizing the visible mask as attention to refine amodal mask (b,c,d) achieves explicitly better performance on amodal mask prediction than using the coarse amodal mask (a). This indicates that applying visible attention for amodal mask prediction is more reasonable than applying amodal attention. Using the amodal mask as attention (c,d) to refine visible mask can get better performance on visible mask prediction than using the coarse visible mask (a,b). This indicates the effect of applying amodal attention in visible mask prediction.  Using the refined visible mask as attention to refine amodal mask (d) has slight improvement than using the coarse visible mask (c). This result shows that using more accurate visible attention can obtain improvement in amodal mask prediction.

\subsection{Visualization of Shape Prior in the Codebook}
We also show some category-specific shape prior clusters such as carrot, bottle and cucumber, via t-SNE \cite{maaten2008visualizing} in Fig. \ref{fig:exp_kmeans}.
For each category, we partition the latent feature into 1024 clusters via K-Means, since we need to use many redundant shape prior items to store the various changes of rotation. 
Thus, there exist a huge number of shape prior in each category in Fig. \ref{fig:exp_kmeans} (a)-(c). 
In particular, Fig. \ref{fig:exp_kmeans} (c) represents the learned codebook of cucumber, just like a circle. As we know, cucumber is a formable object with little shape changes, while the 6 amodal masks reflect its rotation changing process. This visualization shows that the category-specific shape prior has certain interpretability. 

\section{Conclusion}

In this amodal segmentation work, we propose a novel model to mimic the human amodal perception using the shape prior to imagine the invisible regions mainly based on the feature of visible regions. 
However, almost existing methods use the appearance of the whole region-of-interest to infer the amodal masks, which is against the human amodal perception. And this strategy brings the ambiguity that the same appearance of occlusion may require different predictions.
To simulate the imagination from visible region and shape prior, 
we use the visible mask as attention to focus on the visible regions and build a codebook to store the collected amodal shape prior embeddings for refinement and post-process. 
The experimental results indicate our method outperforms other state-of-the-art methods.

\section{Acknowledgment}
The work was supported by National Key R\&D Program of China (2018AAA0100704), NSFC \#61932020,  Science and Technology Commission of Shanghai Municipality (Grant No. 20ZR1436000) and ShanghaiTech-Megvii Joint Lab.
\bibliography{ref}

\end{document}